\documentclass[iclrfinal]{article}

\PassOptionsToPackage{numbers, square, comma}{natbib}

\usepackage{tccml_iclr2025_conference}

\usepackage{ISG_style}
\usepackage{times}  

\usepackage[numbers, square]{natbib} 
\usepackage[utf8]{inputenc} 
\usepackage[T1]{fontenc}    
\usepackage{hyperref}       
\usepackage{url}            
\usepackage{booktabs}       
\usepackage{amsfonts}       
\usepackage{nicefrac}       
\usepackage{microtype}      

\usepackage{graphicx}
\usepackage{multirow}
\usepackage{enumitem}

\usepackage{algorithm}
\usepackage{makecell}
\usepackage{algcompatible}
\usepackage{algpseudocode}

\usepackage{amsmath,amssymb} 
\usepackage{dsfont}

\makeatletter
\NAT@numberstrue  
\makeatother

\DeclareMathOperator*{\argmax}{arg\,max}  

\newcommand{\nn}{{\sf NN}}

\newcommand{\grnn}{{\sf GRNN}}
\newcommand{\gcnn}{{\sf GCNN}}
\newcommand{\gcnns}{{\sf GCNNs}}

\newcommand{\filter}{{\sf H}}

\newcommand{\GRQN}{{\sf GRQN}}
\newcommand{\GRQNs}{{\sf GRQNs}}

\newcommand{\rl}{{\sf RL}}
\newcommand{\PFW}{{\sf PFW}}
\newcommand{\TE}{{\sf TE}}

\newcommand{\episode}{{\sf Episode}}
\newcommand{\buffer}{{\sf Buffer}}
\newcommand{\load}{{\sf load}}
\newcommand{\loss}{{\sf LL}}
\newcommand{\tll}{{\sf TLL}}
\newcommand{\isend}{{\sf end}}
\newcommand{\pf}{{\sf PF}}
\newcommand{\mc}{{\sf MC}}

\title{Real-Time Risky Fault-Chain Search using Time-Varying Graph RNNs}

\author{%
    Anmol~Dwivedi\\
    Rensselaer Polytechnic Institute
    \And
    Ali~Tajer \\
    Rensselaer Polytechnic Institute
}
\iclrfinalcopy

\begin{document}
\maketitle

\begin{abstract}
This paper introduces a data-driven graphical framework for the real-time search of risky cascading fault chains (FCs) in power-grids, crucial for enhancing grid resiliency in the face of climate change. As extreme weather events driven by climate change increase, identifying risky FCs becomes crucial for mitigating cascading failures and ensuring grid stability. However, the complexity of the spatio-temporal dependencies among grid components and the exponential growth of the search space with system size pose significant challenges to modeling and risky FC search. To tackle this, we model the search process as a partially observable Markov decision process (POMDP), which is subsequently solved via a time-varying graph recurrent neural network (GRNN). This approach captures the spatial and temporal structure induced by the system's topology and dynamics, while efficiently summarizing the system's history in the GRNN's latent space, enabling scalable and effective identification of risky FCs.
\end{abstract}

\vspace{-0.2in}

\section{Introduction}
\label{sec:Introduction}

Large-scale disruptions in power-grids often stem from small, unnoticed anomalies that gradually stress the system. Past blackouts show that system operators often miss these signals, allowing failures to cascade into widespread disruptions~\cite{NAPS:2004}. Thus, forming~{real-time} situational awareness is essential to preventing such failures and ensuring a resilient grid operation. This work enhances grid resilience by dynamically predicting the risky fault chains the power system faces, leading to a more secure and reliable grid. A resilient grid is vital for scaling renewable energy integration, facilitating the transition to a low-carbon future, and playing a key role in mitigating climate change.

A fault chain (FC)~\cite{Wang:2011} is a sequence of consecutive component outages that capture the temporal evolution of cascading failures in power-grids. Since a FC captures the high-impact, low-probability nature of cascades, timely identification of the riskiest FCs is crucial for predicting~\cite{Wu:2021, Dobson:2017, Chadaga:2023, Ghosh:2024} and mitigating cascading failures~\cite{Zhou:2020, Yoon:2021, TPS-Blackout-Dwivedi:2024, Neurips-Dwivedi:2024}. Besides, FC search algorithms are widely used for risk assessment~\cite{Rezaei:2015, Henneaux:2016, Rui:2017} and identifying vulnerable grid components~\cite{Wei:2018, Zhang:2020}, among other applications~\cite{Pattern:2021}. Predicting FCs and assessing their risks involves (i) enumerating all possible failure scenarios up to a time horizon, (ii) evaluating the disruption (e.g., load loss) caused by each, and (iii) evaluating the likelihood of each scenario. However, accomplishing these three tasks face the following two key challenges: first, the scenario space grows exponentially with the system size and time horizon, making exhaustively listing all FCs and assessing their risk computationally infeasible, and second, the system's dynamic nature necessitates constantly updating the scenario space.

There are two main approaches for the design of risky FC search algorithms: model-based and data-driven. Model-based methods qualitatively model, analyze and simulate large-scale cascading outage processes by developing simulation models that capture the system physics. For instance, detailed failure models such as the OPA~\cite{Dobson:OPA}, improved OPA~\cite{Improved:OPA}, random chemistry~\cite{RandomChemistryModel} and others~\cite{Soltan:2017, Yang:2017, Ding:2017, Cetinay:2018, Guo:2018} precisely model the AC power-flow and dispatch constraints of the grid components while simulating cascades. Despite their detailed modeling and effectiveness, these models are computationally intensive, a major impediment to their adoption for~\emph{real-time} implementation. The next category of studies develop hybrid high-level statistical models to facilitate faster computation and quick inference such as the CASCADE model~\cite{Dobson:2003}, the branching process model~\cite{Dobson:2004}, the interaction model~\cite{Sun:2015}, and the influence graph models~\cite{Dobson:2017, Wu:2021}. While these approaches are quick in revealing important quantitative and interpretable properties of cascades, such models fail to accommodate the~\emph{time-varying} grid component interactions at different stages of a cascade.

Conversely, machine learning (ML) approaches have been proposed to enhance the efficiency of risky FC search. Broadly, these models assess the vulnerability of each power-grid component using simulated operational data via data-driven algorithms. For instance, the study in~\cite{Yan:2017} formulates the search for risky FCs in a Markov decision process (MDP) environment and employs reinforcement learning (RL) algorithms to find risky FCs. The investigation in~\cite{Li:2018} employs deep neural networks to obtain critical states during cascading outages, and~\cite{Du:2019} employs convolutional neural networks (CNNs) for faster contingency screening. Authors in~\cite{Zhang:2020} propose a transition-extension approach that builds upon~\cite{Yan:2017} to make it amenable to real-time implementation facilitated by exploiting the similarity between adjacent power-flow snapshots and finally, authors in~\cite{Liu:2021} propose to employ graph-CNNs that leverage the grid's topology to identify cascading failure paths.

Despite the effectiveness of the aforementioned ML models, these models broadly face the following challenges: (i) The models fail to capture the concurrent spatio-temporal dependencies across time among the grid components under dynamically~\emph{changing} topologies; (ii) the cascading outage process is assumed to follow Markov property. While this simplification can render reasonable approximations during the earlier stages of a cascade, the later stages of a cascade process typically exhibit temporal dependencies beyond the previous stage, making the Markovian assumptions inadequate; (iii) these models become prohibitive even for moderate grid sizes due to combinatorial growth in either the~\emph{computational} or~\emph{storage} requirements with the number of grid components.

\noindent{\bf Contribution:} We propose a data-driven graphical framework for efficiently identifying risky cascading FCs to address the aforementioned challenges associated with the ML approaches. The proposed framework designs a graph recurrent neural network (GRNN) to circumvent the computational complexities of the real-time search of FCs. The search process is formalized as a partially observable Markov decision process (POMDP), which is subsequently solved via a time-varying GRNN that judiciously accounts for the inherent temporal and spatial structures of the data generated by the system. The key features of the model include (i) leveraging the spatial structure of the data induced by the system topology, (ii) leveraging the temporal structure of data induced by system dynamics, and (iii) efficiently summarizing the system's history in the latent space of the GRNN, rendering the modeling assumptions realistic and the approach amenable to real-time implementation. The data and code required to reproduce our results is publicly~\href{https://github.com/anmold-07/GRNN-FaultChain-Predictor}{available} 

\vspace{-0.1in}

\section{Problem Formulation}
\label{sec:Problem Formulation}

\noindent{\bf Fault Chain Model:} Consider the topology of a generic outage-free system that precedes a FC given by $\mcG_{0}$, and the associated system state by $\bX_{0} \in \R^{N\times F}$ where $F$ and $N$ denotes the number of system state parameters and buses (graph vertices), respectively. We denote the set of all components in the system by $\mcU$. We specify a generic FC that the system might be facing as a~\emph{sequence} of consecutive component outages consisting of at most $P$ stages, where $P$ can be selected based on the horizon of interest for risk assessment. In each stage $i\in[P] \dff [1, \dots, P]$, the set $\mcU_i$ represents the components that fail during stage $i\in[P]$. {Note that the set $\mcU_i\subseteq \mcU\backslash \{\cup_{j=1}^{i-1}\mcU_j\}$ could consist of more than one component failure in any stage $i$, i.e., $|\mcU_i| \geq 1$.
Consequently, we denote the sequence of components that fail during the $P$ stages by a FC~{sequence} $\mcV \dff \langle \mcU_1, \mcU_2, \dots, \mcU_P \rangle$ and denote the healthy components before failure(s) in stage $i$ of the FC sequence by $\ell_{i} \in \mcU\backslash \{\cup_{j=1}^{i-1}\mcU_j\}$.

\noindent{\bf Quantifying Risk:} Due to component outages in each stage $i$, the system's topology alters to $\mcG_{i} \dff (V_{i}, E_{i})$ with an associated adjacency matrix $\bB_{i}$ and an underlying system state $\bX_{i}$. Consecutive failures lead to compounding stress on the remaining grid components, which need to ensure minimal load shedding in the network. Nevertheless, when the failures are severe, they lead to load losses. We denote the load loss $(\loss)$ imposed by the component failures in stage $i$ by $\loss(\mcU_{i}) \dff \load(\mcG_{i-1}) - \load(\mcG_i)$, where $\load(\mcG_{i})$ is the~\emph{total} load (in MWs) when the system's state in stage $i$ is associated with the graph topology $\mcG_i$. Accordingly, we define the~{total} load loss ($\tll$) imposed by any FC sequence $\mcV$ by $\tll(\mcV) \dff \sum_{i=1}^{P} \loss(\mcU_{i})$.

\noindent{\bf Maximizing Accumulated $\tll$:} Due to the re-distribution of power across grid components after each stage $i$ of the FC, some FC sequences particularly lead to substantial risks, and owing to the continuously time-varying system's state $\bX_{0}$, different loading and topological conditions face different risks. Our objective is to identify $S$ number of FC sequences, each consisting of $P$ stages, that impose the~\emph{largest} $\tll$ associated with any given initial observation $\bO_{0} \dff (\mcG_{0}, \bX_{0})$. To formalize this, we define ${\mcF}$ as the set of all possible FC sequences $\mcV$ with a target horizon of $P$, and our objective is to identify $S$ members of~{$\mcF$} with the largest associated risk (load losses). We denote these $S$ members by $\{\mcV^*_1,\dots,\mcV^*_S\}_{\bO_{0}}$. Identifying the sets of interest can be formally cast as solving \begin{align}
\label{eq:OBJ1}
    \mcP:\quad  \{\mcV^*_1,\dots,\mcV^*_S\}_{\bO_{0}} \dff  \argmax_{\{\mcV_1,\dots,\mcV_S\}:\;\mcV_i\in{\mcF}} \;  \sum_{i = 1}^{S} \; \tll(\mcV_{i})\ .
\end{align} Without loss of generality, we assume that the $\tll$s of the set of sequences $\{\mcV^*_1,\dots,\mcV^*_S\}_{\bO_{0}}$ are in the descending order, i.e., $\tll(\mcV^*_1) \geq \tll(\mcV^*_2) \geq \dots \geq \tll(\mcV^*_S)$.

Solving $\mcP$ faces computational challenges since the cardinality of $\mcF$ grows exponentially with the number of components $|\mcU|$ and horizon $P$. To address this, we design an agent-based learning algorithm that~\emph{sequentially} constructs the FCs $\{\mcV^*_1,\dots,\mcV^*_S\}_{\bO_{0}}$. The agent starts by constructing $\mcV^*_1\dff \langle \mcU_{1,1},\dots, \mcU_{1,P} \rangle$ where it sequentially identifies the sets $\{\mcU_{1,i}\}$ in each stage $i$ (via our proposed Algorithm~\ref{alg:ALGO1})} by admitting the observation $\bO_{0}$ as its baseline input. The risk associated with each candidate set $\mcU_{1,i}$ has two, possibly opposing, impacts. The first pertains to the immediate $\loss$ due to component failures in $\mcU_{1,i}$, and the second captures the $\loss$ associated with the future possible failures driven by the failures in $\mcU_{1,i}$. Hence, identifying the sets $\mcU_{1,i}$ involves look-ahead decision-making and cannot be carried out greedily based on only the immediate $\loss$s. 

\noindent{\bf Risky FC Sequential Search as a POMDP:} To control the computational complexity of the search process, each set $\mcU_{1,i}$ is determined from the observation $\bO_{i} = (\mcG_{i}, \bX_{i})$ in the current stage $i$ only. Since the observation $\bO_{i}$ in each stage provides only~\emph{partial} information for decision making despite the $\loss$ at stage $i$ depending on~\emph{all} the past $i$-$1$ stages and the set of components removed in those~stages, we formalize the agent decision process by a partially observed Markov decision process (POMDP) characterized by the tuple $(\mcS, \mcA,\P, \mcR, \mcO, \Z, \gamma)$. Detailed information about the POMDP modeling techniques employed is provided in Appendix~\ref{sec:POMDP Modeling}. After the agent identifies $\mcU_{1,i}$, the agent removes all the components in this set to determine the updated observation $\bO_{i+1} = (\mcG_{i+1}, \bX_{i+1})$ via physical simulation models, using which $\mcU_{1,i+1}$ is determined. This process continues recursively for a total of $P$ stages, at the end of which the set $\mcV^*_1$ is constructed and the algorithm repeats this process $S$ times to identify $S$ FC sequences of interest. The policy $\pi^{*}$ to finding the~\emph{riskiest} FC sequence $\mcV^*_1$ starting from a baseline state $\bS_{0}$~\eqref{eq:POMDP state} can be formally cast as solving $\pi^{*}\dff \argmax_{\pi} \E \left[ V_{\pi}(\bS_{0}) \right]$.

\section{Graphical Risky Fault-Chain Search Framework}
\label{sec:Graphical Risky Fault Chain Search Framework}

\noindent{\bf Motivation:} Model-free off-policy RL algorithms~\cite{sutton:2018} with function approximation, such as deep $Q$-learning~\cite{Mnih:2015}, are effective at finding good policies for high-dimensional state spaces $\mcS$  without requiring access to the transition probability kernel $\P$. However, deep $Q$-learning is ineffective for solving~\eqref{eq:OBJ1} for two reasons: (i) it typically trains a policy to find the riskiest FC $\mcV^*_1$ given a starting state $\bS_0$, whereas we need the $S$ riskiest FCs, and (ii) the partial observations $\bO_i$ at each stage $i$ do not reflect the underlying POMDP state $\bS_i$~\eqref{eq:POMDP state}, leading to $Q(\bO_i, a_i) \neq Q(\bS_i, a_i)$.

\noindent{\bf GRQN Architecture:} To exploit the underlying structure of each POMDP state $\bS_i$~\eqref{eq:POMDP state}, we leverage time-varying graph convolutional filters~\eqref{eq:graph filter} to model the strong correlation induced by the meshed topology of power transmission networks as a latent state $\bZ_{i}\in \R^{N\times H}$. We then model the evolution of this latent state via recurrence~\eqref{eq:GRNN time varying latent state} to account for observational dependencies across stages in the FC sequence $\mcV$. Building on the approach of~\cite{DRQN:2015}, we~\textbf{estimate} $\bY_i \in \mathbb{R}^{N \times G}$~\eqref{eq:GRNN output} from $\bZ_i$~\eqref{eq:GRNN time varying latent state} at each stage $i\in[P]$, and use it to~\textbf{predict} $Q(\bY_i, a_i)$ as a proxy for the $Q$-values of each POMDP state-action pair $Q(\bS_i, a_i)$, assembling into a graph recurrent $Q$-network (GRQN) architecture (more details in Appendix~\ref{sec:GRQN Architecture}).

\noindent{\bf Graph Recurrent $Q$-learning for Risky FC Search:} To identify the $S$ FC sequences with the highest $\tll$ for a given initial observation $\bO_0$, it is insufficient to merely~\textbf{predict} $Q(\bY_i, a_i)$; we must also~\textbf{guide the search} for subsequent FCs with the next highest $\tll$ by leveraging $Q(\bY_i, a_i)$. We propose a graph recurrent $Q$-learning Algorithm~\ref{alg:ALGO1} to sequentially discover the $S$ FC sequences of interest $\{\mcV_1,\dots,\mcV_S\}_{\bO_{0}}$ while concurrently updating the parameters of the $\GRQN$ via~\ref{eq:gradient update}. To avoid repeatedly discovering the same FC sequences, we make two key modifications: first, we alter how ``experience'' is collected in the sequential experience buffer (Appendix~\ref{sec:Sequential Experience Buffer}) by modifying the agent's decision process,~\eqref{eq:exploration policy} and \eqref{eq:exploitation policy}, to track how often each grid component $\ell_{i} \in \mcA$ is removed from each POMDP state $\bS_{i}$ until the current search iteration; and second, instead of re-setting the latent states $\bZ_i$ to zero, we carry forward the previously learned latent state $\bZ_i$ of the GRNN after each newly discovered FC sequence $\mcV_{s}$ during the parameter update~\eqref{eq:gradient update} as the agent gains experience.

\begin{figure*}[t]
    \centering
    \begin{minipage}{0.42\textwidth}
        \centering
        \begin{table}[H] 
            \centering
            \scalebox{0.55}{
                \begin{tabular}{| c | c | c | c | c |} 			
                    \hline
                    \thead{Algorithm}               
                    & \thead{Range for Accumulated $\tll$\\ $\sum_{i=1}^{S} \tll(\mcV_{i})$ (in GWs)}   
                    & \thead{Range for} ${\sf Regret}(S)$ (in GWs) \\ [2.0ex]
                    \hline
                    Algorithm~\ref{alg:ALGO1}$\;(\kappa = 3)$    & $110.42\;\pm\;37\%$ & $765.33\;\pm\;5.3\%$      \\ [1.0ex] 
                    \hline
                    Algorithm~\ref{alg:ALGO1}$\;(\kappa = 2)$    & $96.18\;\pm\;38\%$  & $779.57\;\pm\;4.7\%$     \\ [1.0ex] 
                    \hline
                    Algorithm~\ref{alg:ALGO1}$\;(\kappa = 1)$    & $86.43\;\pm\;35\%$  & $789.32\;\pm\;3.87\%$     \\ [1.0ex] 
                    \hline
                    $\PFW$ + $\rl$ + $\TE$ \cite{Zhang:2020} & $60.63\;\pm\;3.4\%$  & $815.12\;\pm\;0.26\%$    \\[1.0ex] 
                    \hline
                    $\PFW$ + $\rl$ \cite{Zhang:2020}  & $57.18\;\pm\;3.1\%$  & $818.57\;\pm\;0.22\%$   \\ [1.0ex] 
                    \hline
                \end{tabular}%
            }
            \caption{Performance comparison for the IEEE-$39$ bus system under unbounded computational budget.}
            \label{table:39 bus}
        \end{table}
    \end{minipage}
    \hfill
    \begin{minipage}{0.51\textwidth}
        \centering
        \includegraphics[width=\linewidth]{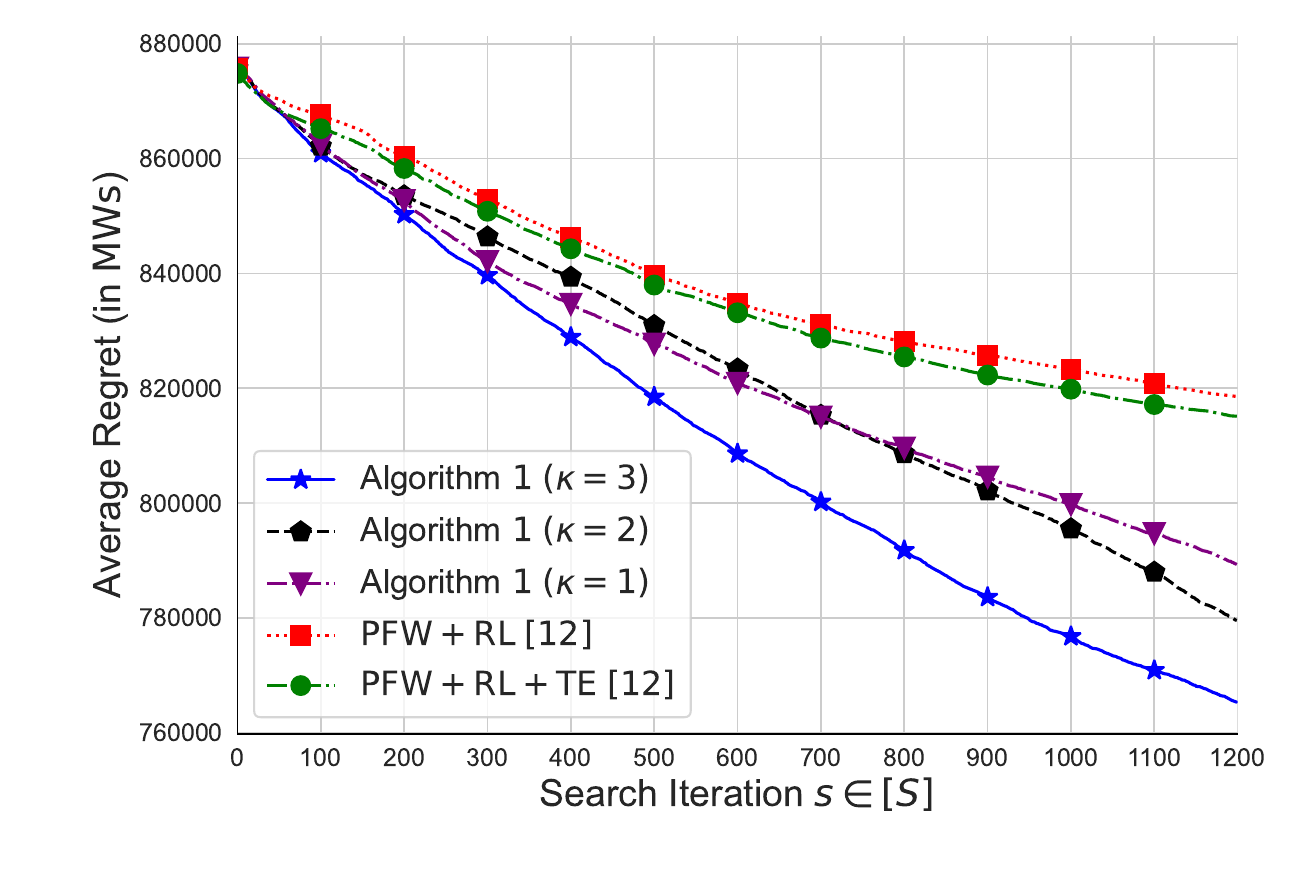}
        \vspace{-0.4in} 
        \caption{${\sf Regret}(s)$ versus $s$.}
        \label{fig:Accum Risk}
    \end{minipage}
\end{figure*}

\vspace{-0.08in}

\section{Experiments}
\label{sec:Experiments}

\noindent{\bf Comparison under Unbounded Computational Budget:} We compare the accumulated $\tll$ from the predicted FCs $\{\mcV_1,\dots,\mcV_S\}_{\bO_{0}}$ in $S=1200$ search iterations and benchmark the accuracy of Algorithm~\ref{alg:ALGO1} against \textbf{ground-truth} by generating optimal FCs for the IEEE 39- and 118-bus systems. We calculate~\textbf{regret} that quantifies the gap between the accumulated $\tll$ from the~\textbf{optimal} FCs $\{\mcV^*_1,\dots,\mcV^*_S\}_{\bO_{0}}$ and the predicted FCs. Until iteration $s \in [S]$, we define ${\sf Regret}(s) \dff \sum_{i=1}^{S} \tll(\mcV^{*}_{i}) - \sum_{i=1}^{s} \tll(\mcV_{i})$. Lower regret indicates a higher accuracy. Detailed descriptions of the algorithm, parameters, and baselines are in Appendix~\ref{sec:Architecture, Training, Initialization},~\ref{sec:IEEE-39 New England Test System} and~\ref{sec:Baseline Agents}. Alongside Algorithm~\ref{alg:ALGO1}, we run $Q$-learning updates outlined in~\cite{Zhang:2020} for both $\PFW$+$\rl$ and $\PFW$+$\rl$+$\TE$ baseline agents. Table~\ref{table:39 bus} compares the accumulated $\tll$ and regret (mean $\pm$ SD) for various agents with~\textbf{decreasing} computational complexity as we move down the table. Algorithm~\ref{alg:ALGO1} consistently outperforms the baseline approaches, with $\kappa=3$~\textbf{reducing} average regret by $6.2\%$ compared to the best baseline ($\PFW$ + $\rl$ + $\TE$) and achieving almost~\textbf{double} the accumulated $\tll$. We also observe that a greater $\kappa$ results in~\textbf{larger} accumulated $\tll$ and~\textbf{lower} regret on average since the weights of the behavior $\GRQN$ are updated more frequently (Algorithm~\ref{alg:ALGO1}), leading to better $Q$-value predictions for each~{POMDP} state $\bS_{i}$. For e.g., the average regret of Algorithm~\ref{alg:ALGO1} with $\kappa = 3$ is $765.3$ GW, $3.04\%$ lower than when $\kappa = 1$. Figure\ref{fig:Accum Risk} further shows regret as a function of search iterations $s\in[S]$. 

\noindent{\bf Comparison under Bounded Computational Budget:} We next evaluate the performance of the agents within a strict~\textbf{5-minute} computational time budget, simulating real-time implementation since the complexity for FC search should be within the dispatch cycle. Table~\ref{table:time_39 bus} summarizes the results and three key observations emerge: (i) With a 5-minute budget, Algorithm~1 with $\kappa = 3$ discovers~\textbf{fewer} FC sequences compared to other algorithms. This is expected, as a higher $\kappa$ requires more computation per iteration (Algorithm~\ref{alg:ALGO1}), reducing the total number of iterations and hence, discovered FC sequences $S$; (ii) Algorithm~\ref{alg:ALGO1} with $\kappa = 2$ finds the~\textbf{most} accumulated $\tll$. Although $\PFW$ + $\rl$ and $\PFW$ + $\rl$ + $\TE$ discover more FC sequences (1611 and 1608, respectively), their accumulated $\tll$ are lower than Algorithm~\ref{alg:ALGO1} with $\kappa = 2$; (iii) In terms of accuracy, Algorithm~\ref{alg:ALGO1} with $\kappa = 3$ outperforms other algorithms, despite finding fewer FC sequences (575). The frequent weight updates of the $\GRQN$ lead to more accurate $Q$-value predictions, better~\textbf{optimizing performance per search iteration}. 

\noindent{\bf Discussion:} It is noteworthy that all agents were evaluated on a standard computer~\textbf{without} GPUs. Leveraging GPUs will further accelerate the search process for Algorithm~\ref{alg:ALGO1}, as they are designed for matrix and vector operations, unlike the baseline approaches, which cannot be similarly accelerated. We also emphasize that the baseline approach~\cite{Zhang:2020} models each~\emph{permutation} of component outages as a unique MDP state, and stores the $Q$-values for a combinatorial number of resulting~{MDP} state-action pairs in an extensive $Q$-table, rendering it not scalable. However, by judiciously leveraging the graphical structure of each POMDP state and appropriately modeling the dependencies across the various stages of the FC, we have bypassed the storage challenge with fewer modeling assumptions while, at the same time, achieving better performance. Similar trends for the IEEE 118-bus system is provided in Appendix~\ref{sec:IEEE-118 test system}, confirming the trends observed for the IEEE 39-bus system.

\newpage

\begin{table*}[t]
	\centering
	\scalebox{0.8}{
		\begin{tabular}{| c | c | c | c | c | c |} 			
			\hline
			\thead{Algorithm}     
			& \thead{Average No. of FC \\ Sequences $S$ Discovered}
            & \thead{Range for Accumulated $\tll$\\ $\sum_{i=1}^{S} \tll(\mcV_{i})$ (in GWs)}   
			& \thead{Range for} ${\sf Regret}(S)$ (in GWs) \\ [2.0ex] 
			\hline
			
			Algorithm~\ref{alg:ALGO1}$\;(\kappa = 3)$  & $575$  & $63.484\;\pm\;36.7\%$      & $457.09\;\pm\;5.9\%$    \\ [1.0ex] 
			\hline
			
			Algorithm~\ref{alg:ALGO1}$\;(\kappa = 2)$  & $700$  & $79.792\;\pm\;39\%$    & $521.47\;\pm\;7.5\%$     \\ [1.0ex] 
			\hline
			
			Algorithm~\ref{alg:ALGO1}$\;(\kappa = 1)$  & $937$  & $70.848\;\pm\;33.3\%$   & $673.56\;\pm\;4.76\%$    \\ [1.0ex] 
			\hline
			
			$\PFW$ + $\rl$ + $\TE$ \cite{Zhang:2020} & $1608$ & $68.74\;\pm\;3.4\%$   & $965.41\;\pm\;1.61\%$   \\[1.0ex] 
			\hline
			
			$\PFW$ + $\rl$ \cite{Zhang:2020} & $1611$  & $65.35\;\pm\;3.9\%$   & $969.78\;\pm\;2.27\%$  \\ [1.0ex] 
			\hline
		\end{tabular}%
	}
	\caption{{Performance comparison for a computational time of $5$ minutes for the IEEE-$39$ bus system.}}
	\label{table:time_39 bus}
\end{table*}

\section{Conclusion}
\label{sec:Conclusion}

In this paper, we have considered the problem of real-time risky fault chain identification in a limited number of search trials. We have proposed a data-driven graphical framework that can dynamically predict the chains of risky faults a power system faces. First, the search for risky fault chains is modeled as a partially observed Markov decision process (POMDP). Then a graph recurrent $Q$-learning algorithm is designed to leverage the grid's topology to discover new risky fault chains efficiently. Experimental results on IEEE standard systems, compared to baseline methods, demonstrate the effectiveness and efficiency of our approach.

\bibliography{GRNN_ICLR_2025}

\appendix

\newpage 

\section{Appendix}
\label{sec:Appendix}

\subsection{POMDP Modeling}
\label{sec:POMDP Modeling}

\paragraph{State Space $\mcS$:} We denote the partial observation that the agent uses at stage~$i$ to determine the system state at stage $i+1$ by $\bO_i \dff (\mcG_i, \bX_i)$. Accordingly, we define the sequence
\begin{align}
    \label{eq:POMDP state}
    \bS_{i} \dff \langle \bO_{0}, \dots, \bO_{i} \rangle\ ,
\end{align} which we refer to as the POMDP state at stage $i$, and it characterizes the entire past sequence of observations that render $\bO_{i+1}$. As stated earlier, at stage $i$ only $\bO_i$ is known to the agent. 

\paragraph{Action Space $\mcA$:} At stage $i$, upon receiving the observation $\bO_{i}$, the agent aims to choose a component from the set of~\emph{available} components to be removed in the next stage. To formalize this process, we define the agent's action as its choice of the component of interest. We denote the action at stage $i$ by $a_i$. Accordingly, we define the action space $\mcA_{i}$ as the set of all remaining (healthy) components, i.e., $\mcA_{i} \dff \mcU\backslash \{\cup_{j=1}^{i-1}\mcU_j\}$. It is important to note that to reflect the reality of FCs, in which failures occur component-by-component, we are interested in identifying only~\emph{one} component in each stage $i\in[P]$. Nevertheless, due to the physical constraints, removing one component can possibly cause outages in one or more other components in the~\emph{same} stage. Hence, we denote the set of all components to be removed in stage $i$ by the set $\mcU_{i}$.

\paragraph{Transition Kernel $\P$:} Once the agent takes an action $a_{i} \in \mcA_{i}$ in stage $i$, the underlying POMDP state in the next stage is randomly drawn from a transition probability distribution $\P$
\begin{align}
    \label{eq:system dynamics}
    \bS_{i+1} \sim \P(\bS \;|\;\bS_{i}, a_{i})\ .
\end{align} The probability distribution $\P$ captures the randomness due to the power system dynamics, and it is determined by the generator re-dispatch strategy employed in each stage of the FC. Nevertheless, for a given generator re-dispatch strategy and a given initial state distribution $\bS_{0} = \bO_{0}$ for which the $S$ FC sequences of interest are to be determined, the transitions kernel is~\emph{deterministic}. 

\begin{figure}[t]
	\centering
	\includegraphics[width=0.7\linewidth]{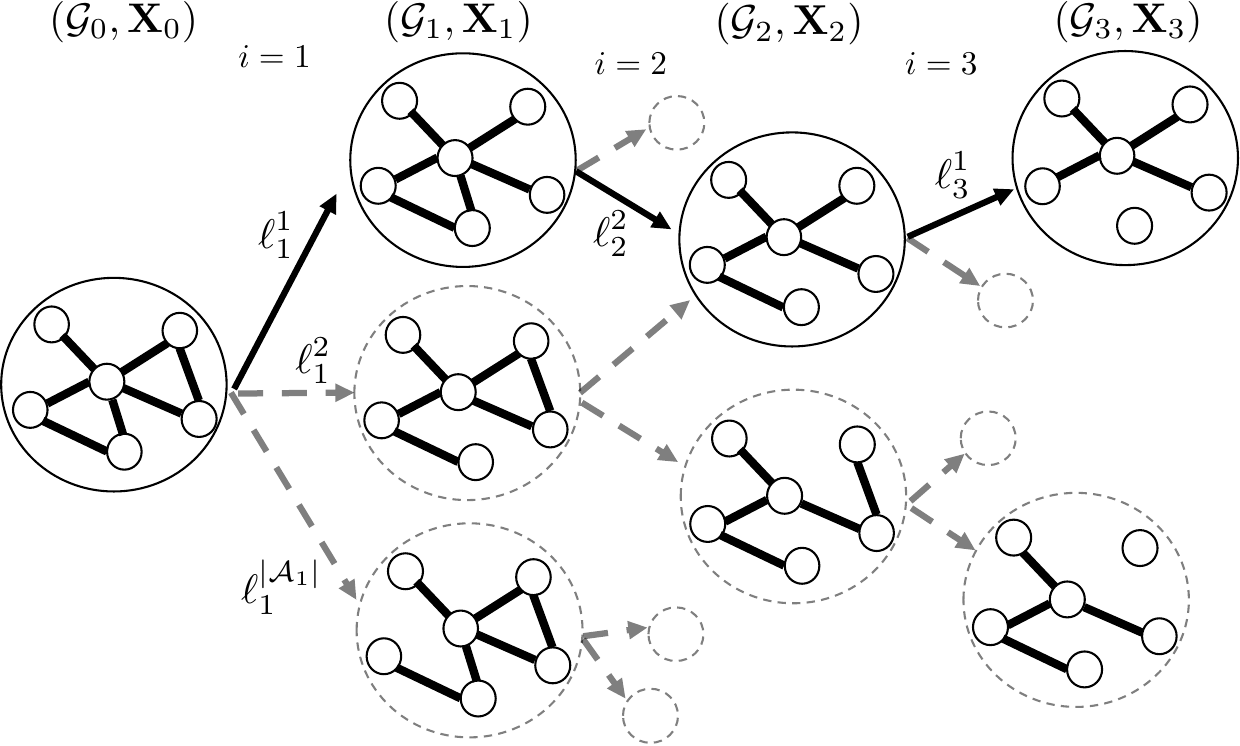}
	\caption{An agent decision process rendering a FC $\mcV^*_{s} = \langle \ell^{1}_{1}, \ell^{2}_{2}, \ell^{1}_{3}\rangle$.}
	\label{fig:look ahead search}
\end{figure}

\paragraph{Reward Dynamics $\mcR$:} To quantify the risk associated with taking action $a_i$ in POMDP state $\bS_{i}$ when transitioning to POMDP state $\bS_{i+1}$, we define an instant reward $r_i$
\begin{align}
\label{eq:rewards}
r_i \dff  r(\bS_{i+1}\;|\; \bS_{i}, a_{i}) \dff \load(\mcG_{i}) - \load(\mcG_{i+1})\ .
\end{align} Hence, for any generic action selection strategy $\pi$, the aggregate reward collected by the agent starting from the baseline POMDP state $\bS_{0}$ can be characterized by a value function 
\begin{align}
    \label{eq:value function}
    V_{\pi}(\bS_{0}) \dff \sum_{i=0}^{P-1} \; \gamma^{i} \cdot r(\bS_{i+1}\;|\; \bS_{i}, \pi(\bO_{i}))\ ,
\end{align} where the discount factor $\gamma\in\R_+$ decides how much future rewards are favored over instant rewards, and $\pi(\bO_{i})$ denotes the action selected by the agent given an observation $\bO_{i}$ in stage $i \in [P]$. Fig.~\ref{fig:look ahead search} illustrates a search process where an agent constructs a FC sequence $\mcV^*_{s} = \langle \ell^{1}_{1}, \ell^{2}_{2}, \ell^{1}_{3}\rangle$ by leveraging the current system state $(\mcG_{i}, \bX_{i})$ in each stage $i\in [3]$.

\subsection{Time-Varying GRNN Model}
\label{sec:Time-Varying GRNN Model}

GRNNs are a family of GNN architectures~\cite{Scarselli:2008} specialized for processing sequential graph structured data streams~\cite{Ruiz:2020}. These architectures exploit the~\emph{local} connectivity structure of the underlying network topology $\mcG_i$ to efficiently extract features from each bus by sequentially processing time-varying system states $\bX_i$ that evolve on a sequence of graphs $\mcG_{i}$. More broadly, these architectures generalize recurrent neural networks~\cite{recurrent:2001} to graphs. To lay the context for discussions, we first discuss time-varying graph convolutional neural networks $(\gcnns)$, the components of which serve as an essential building block to motivate time-varying GRNNs.

\paragraph{Time-Varying GCNNs:} Consider the $i^{\rm th}$ stage of a FC sequence $\mcV_{s}$ where an agent receives an observation $\bO_i \dff (\mcG_i, \bX_i)$ on graph $\mcG_i$ associated with an adjacency matrix $\bB_{i}$. Central to the development of GCNNs is the concept of a~\emph{graph-shift} operation that relates an input system state $\bX_{i}\in \R^{N\times F}$ to an output system state $\bF_{i} \in \R^{N\times F}$ 
\begin{align}
    \label{eq:graph shift}
    \bF_{i} \dff \bB_{i} \cdot \bX_{i}\ .
\end{align} Clearly, the output system state $\bF_{i}$ is a locally shifted version of the input system state $\bX_{i}$ since each element $[\bF_{i}]_{u,f}$, for any bus $u\in V_{i}$ and parameter~$f$, is a linear combination of input system states in its $1$-hop neighborhood. Local operations, such as~\eqref{eq:graph shift} capture the $1$-hop structural information from $\bX_{i}$ by~\emph{estimating} another system state $\bF_{i}$ and are important because of the strong spatial correlation that exists between $[\bX_{i}]_{u,f}$ and its network neighborhood determined by $\mcG_i$. Alternative transformations such as that employed in~\cite{Dwivedi:2022} quantify the merits of estimating system states that capture the spatial structure of $\bX_i$. In order to capture the structural information from a broader $K$-hop neighborhood instead,~\eqref{eq:graph shift} can be readily extended by defining a time-varying~\emph{graph convolutional filter} function $\sf{H}$ $: \R^{N \times F} \rightarrow \R^{N \times H}$ that operates on an input system state $\bX_i$ to~\emph{estimate} an output system state $\filter(\mcG_i, \bX_i\;;\;\mcH)$
\begin{align}
    \label{eq:graph filter}
    \filter(\mcG_i, \bX_i\;;\;\mcH) \dff \sum_{k=1}^{K} \; \left[\bB_{i}^{k-1} \cdot \bX_{i}\right] \cdot \bH_{k}\ ,
\end{align} 
where $H$ denotes the number of output features~\emph{estimated} on each bus and $\mcH \dff \{\bH_{k} \in \R^{F \times H}: k \in [K]\}$ denotes the set of filter coefficients parameterized by matrices $\bH_{k}$~\emph{learned} from simulations where each coordinate of $\bH_{k}$ suitably weighs the aggregated system state obtained after $k$ repeated $1$-hop graph-shift operations performed on $\bX_{i}$. Note that~\eqref{eq:graph filter} belongs to a broad family of~\emph{graph-time filters}~\cite{Isufi:2017} that are polynomials in time-varying adjacency matrices $\bB_{i}$. 
There exists many types of~\emph{graph-time filters}~\cite{Gama:2021}. We employ~\eqref{eq:graph filter} due to its simplicity. Nevertheless,~\eqref{eq:graph filter} only captures simple linear dependencies within $\bX_{i}$. To capture non-linear relationships within $\bX_{i}$, time-varying GCNNs~\emph{compose} multiple layers of graph-time filters~\eqref{eq:graph filter} and non-linearities such that the output system state of each $\gcnn$ layer is given by 
\begin{align}
    \label{eq:gcnns}
    \Phi(\mcG_i, \bX_{i}\;;\;\mcH) \dff \sigma\left(\;\filter(\mcG_i, \bX_{i}\;;\;\mcH)\;\right)\  ,
\end{align} where $\sigma:\R \rightarrow \R$ is commonly known as the activation function (applied element-wise) such that $\Phi(\mcG_i, \bX_{i}\;;\;\mcH) \in \R^{N\times H}$.

\paragraph{Time-Varying GRNNs:} GCNNs can only extract spatial features from each system state $\bX_{i}$ independently~\eqref{eq:gcnns}. For this reason, we add recurrency to our $\gcnn$ model~\eqref{eq:gcnns} in order to capture the temporal observational dependencies across the various stages of a FC sequence to construct a GRNN. A time-varying GRNN extracts temporal features from an input sequence $\langle \bO_i:i \in [P]\rangle$ by~\emph{estimating} a sequence of latent system states $\langle\bZ_{i}:i \in [P]\rangle$ where each~{system} state $\bZ_{i}\in \R^{N\times H}$ is latent that facilitates in summarizing the entire past observational history, that is both redundant and difficult to store, until stage $i$. This is done by judiciously parameterizing each latent system state $\bZ_{i}$ that is a function of the graph-time filter output~\eqref{eq:graph filter} operated on both the current input system state $\bX_i$ and previous latent system state $\bZ_{i-1}$ independently to obtain
\begin{align}
    \label{eq:GRNN time varying latent state}
    \bZ_{i} \dff \sigma\left(\;\filter_{1}(\mcG_i, \bX_{i}\;;\;\mcH_{1}) \; + \; \filter_{2}(\mcG_{i-1}, \bZ_{i-1}\;;\;\mcH_{2}) + b_{\bZ} \;\right)\ ,
\end{align} where $\filter_{1}:\R^{N\times F} \rightarrow \R^{N\times H}$ and $\filter_{2}:\R^{N\times H} \rightarrow \R^{N \times H}$ are filters~\eqref{eq:graph filter} each parameterized by a distinct set of filter coefficients $\mcH_{1}=\{\bH^{1}_{k} \in \R^{F \times H}\; \forall\;k \in [K]\}$ and $\mcH_{2}=\{\bH^{2}_{k} \in \R^{H \times H}\; \forall\;k \in [K]\}$, respectively. Subsequently, similar to~\eqref{eq:gcnns}, to capture the non-linear relationships from each latent system state $\bZ_{i}$ to facilitate dynamic decision-making on graphs $\mcG_i$, the~\emph{estimated} output system state $\bY_{i}\in \R^{N \times G}$
\begin{align}
    \label{eq:GRNN output}
    \bY_{i} \dff \rho\left(\; \filter_{3}(\mcG_i, \bZ_{i}\;;\; \mcH_{3}) + b_{\bY} \;\right) \quad \forall i \in [P]\ ,
\end{align} where the graph-time filter $\filter_{3}:\R^{N\times H} \rightarrow \R^{N \times G}$ in~\eqref{eq:graph filter} is parameterized by the filter coefficient set $\mcH_{3}$, $G$ denotes the number of~\emph{estimated} output features on each bus $u\in V_{i}$, and $\rho$ is a pointwise non-linearity applied element-wise to output $\filter_{3}$. Note that $H$, $K$, $G$, $\sigma$ and $\rho$ are hyper-parameters for a GRNN. Additionally, the number of~\emph{learnable} parameters $\mcH_{i} \; \forall i \in [3]$ is independent of the system size $N$ and the horizon $P$ of the FC sequence due to parameter sharing across the stages of the FC, providing the model with flexibility to~\emph{learn} from input sequences $\langle \bO_i:i \in [P]\rangle$ of different and long risk assessment horizons without a combinatorial growth in the number of~\emph{learnable} parameters, ensuring tractability. Next, we leverage the sequence of GRNN output system states $\langle \bY_{i}:i\in [P]\rangle$ to find the $S$ risky FCs~\eqref{eq:OBJ1} of interest.

\begin{figure}[t]
    \centering
    \includegraphics[width=0.75\linewidth]{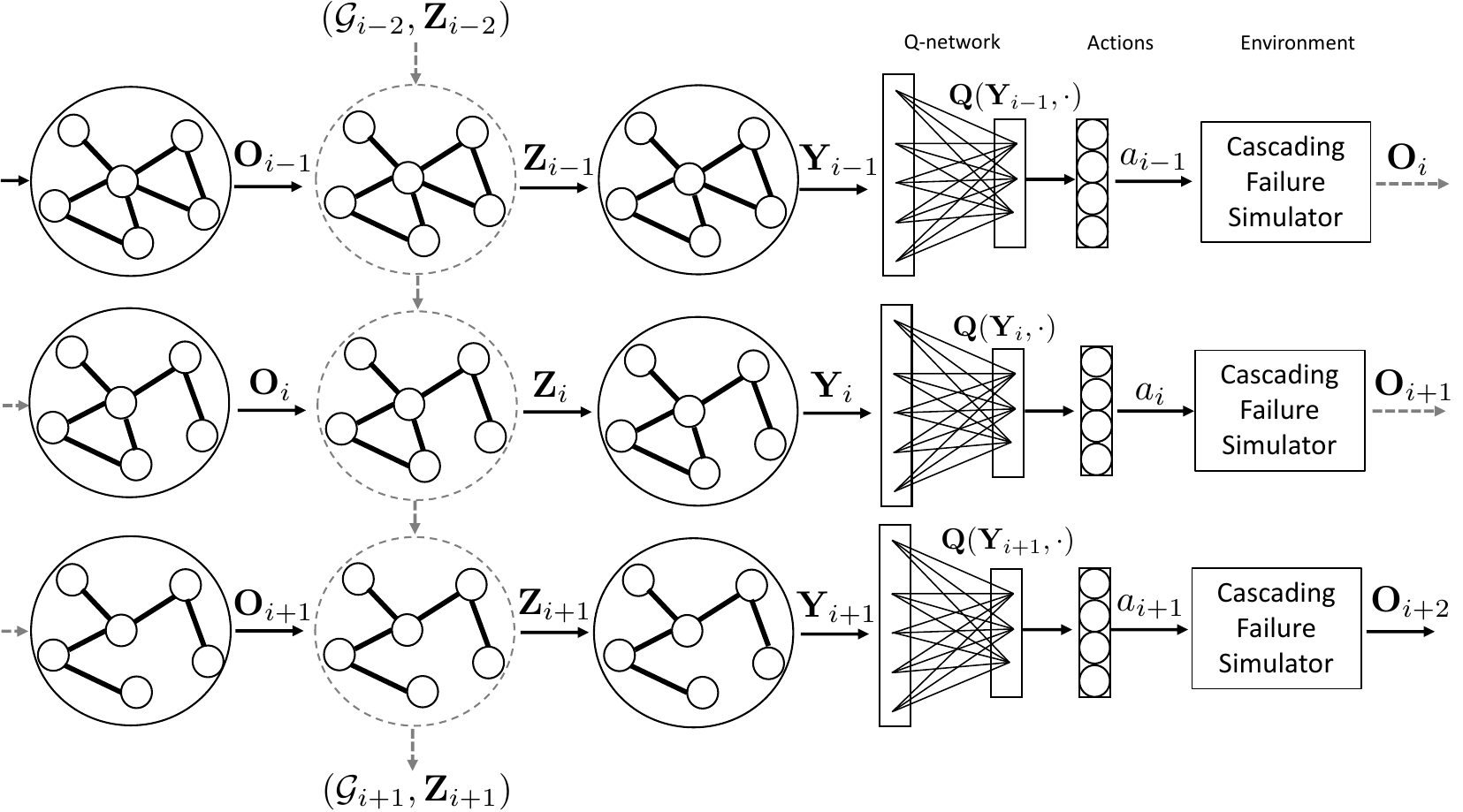}
    \caption{Graph recurrent $Q$-network $(\GRQN)$ architecture.}
    \label{fig:GRQN arch}
\end{figure}

\subsection{Architecture, Training, Algorithm Initialization} 
\label{sec:Architecture, Training, Initialization}

\subsubsection{$\GRQN$ Architecture} 
\label{sec:GRQN Architecture}

The architecture consists of a time-varying GRNN, parameterized by $\btheta_{\grnn} \dff \{\mcH_{i} : i \in [3]\}$, that sequentially processes observations at each stage of the FC and a fully-connected $\nn$, parameterized by $\btheta_{\nn}$, to~\emph{predict} $Q(\bY_{i}, a_{i})$ for each action $a_{i}\in \mcA_{i}$. Accordingly, we parameterize the $\GRQN$ by $\btheta \dff \{\btheta_{\grnn}, \btheta_{\nn}\}$. For any FC sequence $\mcV_{s}$, an input data stream of observations $\bO_{i}\in \R^{N \times F}$ obtained at each stage $i$ of the FC acts as an input to the GRNN using which an output system state $\bY_{i} \in \R^{N \times G}$ is estimated. Subsequently, a fully connected $\nn$ of input and output dimension $N \times G$ and $|\mcU|$, respectively, is leveraged to output a $\bQ(\bY_{i}, \cdot | \btheta) \in \R^{|\mcU|}$ vector consisting of $Q$-values for each action. Overall, at each stage $i$, a $\GRQN$ takes $\bO_{i}$ and $\bZ_{i-1}$ as it's baseline inputs, concisely denoted by $\GRQN(\bO_{i}, \bZ_{i-1}| \btheta)$, and outputs a vector $\bQ(\bY_{i}, \cdot|\btheta)$ and the next latent~{system} state $\bZ_{i}$~\eqref{eq:GRNN time varying latent state}, crucial to carry forward to guide the search of FCs during the training of the $\GRQN$. Note that we initialize $\bZ_{0}$ by $\mathbf{0} \in \R^{N \times H}$. Fig.~\ref{fig:GRQN arch} illustrates the end-to-end $\GRQN$ architecture.

\subsubsection{Sequential Experience Buffer} 
\label{sec:Sequential Experience Buffer}

In order to deal with the issue of~\emph{catastrophic forgetting}~\cite{Mnih:2015}, we employ a~\emph{sequential} experience buffer that stores FC sequences discovered during training of the $\GRQN$ from which batches of random sequences are sampled to facilitate the learning of parameters $\btheta$. While there exists various ways to implement such a buffer, we employ an ordered list that stores all the visited transition tuples as a sequence $\langle (\bO_{i}, a_{i}, r_{i}, \bO_{i+1}, \isend(\bO_{i+1})):i \in [P]\rangle$ where we have defined $\isend(\bO_{i+1})$ as a boolean value, if true, indicating that the observation $\bO_{i+1}$ is associated with a last stage of the risk assessment horizon $P$.

\subsubsection{Training the $\GRQN$} 
\label{sec:Training the GRQN}

For stability in the training of the $\GRQN$, we employ the standard trick~\cite{Mnih:2015} of splitting the task of predicting and evaluating $Q$-values via two separate $\GRQNs$, a target network $\GRQN(\cdot, \cdot|\btheta^{-})$ and a behavior network $\GRQN(\cdot, \cdot|\btheta)$ each parameterized by a distinct set of parameters $\btheta^{-}$ and $\btheta$, respectively. For every training iteration $n$ of the graph recurrent $Q$-learning algorithm, the agent samples $B$ random batches of FC sequences, each of type $\langle (\bO_{j}, a_{j}, r_{j}, \bO_{j+1}, \isend(\bO_{j+1})):j \in [P]\rangle$, from the sequential experience buffer on which the target $\GRQN$ is~\emph{unrolled} to estimate $\bY_j$~\eqref{eq:GRNN output} using which $B$ batches of $\bQ(\bY_j, \cdot|\btheta^{-})$ are predicted with respect to the target network. Subsequently, the agent computes a~\emph{look-ahead} target output for each batch where each target $t_{j} \; \forall j \in [P]$ is given by
\begin{align}
    \label{eq:target update}
    t_{j} = r_{j} +  \gamma \cdot (1 - \isend(\bO_{j+1})) \cdot \max_{a}Q(\bY_{j+1}, a | \btheta^{-})\ .
\end{align} Accordingly, the parameters of the behavior  network is updated via gradient descent with respect to a quadratic loss
\begin{align}
    \label{eq:gradient update}
    \btheta_{n+1} & = \btheta_{n} - \alpha \cdot \nabla_{\btheta_{}}(t_{j} - Q(\bY_{j}, a_{j} | \btheta_{}))^{2}\ , 
\end{align} where $\alpha$ denotes the learning rate and $n$ denotes the current training iteration. The update~\eqref{eq:gradient update} is preformed $\kappa$ times for every action taken by the agent in any stage $i\in[P]$ of the FC and additionally, serves as a means to control the~\emph{computational complexity} of our learning algorithm. In this paper, we employ the Adam optimizer~\cite{adam:2015} to perform the gradient update~\eqref{eq:gradient update} and update the target network parameters $\btheta^{-} = \btheta$ at the end of every FC sequence discovered. 

\subsubsection{Graph Recurrent latent~{System} State Updates} As the agent gains more experience and continues to store visited transition sequences of tuples in the experience buffer, the latent system state $\bZ_{i}$ of the GRNN may either be zeroed or carried forward after every newly discovered FC. Our experiments suggest that sequential updates where the latent~{system} state $\bZ_{i}$~\eqref{eq:GRNN time varying latent state} is carried forward from the previous stages throughout the parameter update~\eqref{eq:gradient update} leads to learning of better FC search strategies. Hence, we choose to carry forward the previously learned latent system state during the training of our $\GRQN$.

\setlength{\textfloatsep}{0pt}
\begin{algorithm}[t]
\caption{Graph Recurrent $Q$-learning ($\GRQN$)}
\label{alg:ALGO1}
\begin{algorithmic}[1]
    \footnotesize
    \Procedure{Graph Recurrent $Q$-learning}{} 
        \State Initialize behaviour  network with random $\btheta$ $\GRQN(\cdot, \cdot | \btheta)$
        \State Initialize target network with weights $\btheta^{-} = \btheta$ $\GRQN(\cdot, \cdot | \btheta^{-})$ 	
        \State Initialize $\sf{Buffer} \leftarrow \langle \rangle$ from Appendix~\ref{sec:Sequential Experience Buffer}
        \State Initialize $\bZ_{0} \leftarrow \mathbf{0} \in \R^{N\times H}$ 	
        \For {$\episode \; s = 1, \dots, S$} 
            \State $\episode \leftarrow \langle \rangle$  
            \State $\mcV_{s} \leftarrow \langle \rangle$
            \State Reset power-flow according to initial state $\bS_{0}, \bO_{0}$
            \For {Stage $i = 1, \dots, P$} 
                \State $\_\_, \bZ_{i} \leftarrow \GRQN(\bO_{i}, \bZ_{i-1} | \btheta)$
                \State $a_{i} \leftarrow 
                \begin{cases}
                    \mbox{explore via}~\eqref{eq:exploration policy}  & \text{if rand}(0, 1) \leq \epsilon \\
                    \mbox{exploit via}~\eqref{eq:exploitation policy} & \text{otherwise}
                \end{cases}$
                \State Take action $a_{i}$ and determine $\mcU_{i}$
                \State $\mcV_{s} \leftarrow  \mcV_{s} \cup \mcU_{i}$
                \State Update power-flow and obtain $\bO_{i+1}$
                \State Calculate load loss $r_{i}$ from~\eqref{eq:rewards}
                \State $\episode \leftarrow \episode \cup (\bO_{i}, a_{i}, r_{i}, \bO_{i+1}, \isend(\bO_{i+1}))$
                \If{$s \geq {\sf Explore}$}
                    \State ${\sf count}(\bS_{i},a_{i}) \leftarrow {\sf count}(\bS_{i},a_{i}) + 1$ 
                    \For {training $n = 1, \dots, \kappa$} 
                        \State Sample $B$ FC sequences from $\buffer$
                        \State $t_{j}, \_\_ \leftarrow \GRQN(\bO_{j+1}, \mathbf{0} | \btheta^{-})$ from~\eqref{eq:target update}
                        \State $Q(\bY_{j}, a_{j}| \btheta), \_\_  \leftarrow \GRQN(\bO_{j}, \mathbf{0} | \btheta)$
                        \State Calculate $\nabla_{\btheta_{}}(t_{j} - Q(\bY_{j}, a_{j} | \btheta))^{2}$ 
                        \State Update $\btheta$ as in~\eqref{eq:gradient update}
                        \State Update $\epsilon$ as in~\eqref{eq:epsilon update} 
                    \EndFor	
                \EndIf
                \State Update availability of actions backwards
            \EndFor	
            \State $\sf{Buffer} \leftarrow \sf{Buffer} \cup \episode$
            \State $\bZ_{0} \leftarrow \bZ_{P}$
            \If{$\tll(\mcV_{s}) \geq M$}
                \State Store risky FC
            \EndIf
            \State $\btheta^{-} = \btheta$
        \EndFor
    \EndProcedure
    \State Find the accumulated risk due to all the $S$ FCs $\{\mcV_1,\dots,\mcV_S\}$.
\end{algorithmic}
\end{algorithm}

\subsubsection{Outline of Algorithm 1} 
\label{sec:Outline of Algorithm 1}

Algorithm~\ref{alg:ALGO1} outlines the graph recurrent $Q$-learning algorithm used to~\emph{learn} the parameters $\btheta$ of the behaviour $\GRQN(\cdot, \cdot | \btheta)$ to facilitate the discovery of the $S$ {number of} FC sequences of interest. During the initial few iterations, since the sequential experience buffer is empty, we allow the agent to~\emph{explore} and fill the buffer with FC sequences for ${\sf Explore}$ number of iterations~\emph{offline} (more details in Section~\ref{sec:Offline Sequential Buffer Initialization}). Subsequently, the real-time algorithm is initiated. Initially, the agent lacks any information about the cascading failure dynamics. Hence, it relies on the~\emph{prior knowledge} to construct~$\mcU_{j,i}$, for any fault chain $j$ in each stage $i\in [P]$, as the agent is unable to evaluate the $\loss$ associated with removing an arbitrary component $\ell_{i} \in \mcU_j \backslash \{\cup_{k=1}^{i-1}\mcU_{j, k}\}$ in any stage of the FC $\mcV_{j}$. Gradually, Algorithm~\ref{alg:ALGO1} learns to construct the sets~$\mcU_{j,i}$ by leveraging the learned latent graphical feature representations. These representations are obtained by optimizing the~\emph{look-ahead} target function~\eqref{eq:target update} characterized by the behavior network $\GRQN(\cdot, \cdot|\btheta)$ since the parameters $\btheta$ of this network determine the $Q$-values influencing the actions $a_i \in \mcA_i$ taken by the agent. Therefore, when choosing actions $a_i \in \mcA_i$, the agent should make a trade-off between~\emph{exploration} and~\emph{exploitation} throughout the training of the $\GRQNs$. Next, we discuss an~\emph{exploration-exploitation} search strategy that the agent employs to make the real-time search of FCs of interest more efficient. 

\subsubsection{Fault Chain Search Strategy}
\label{sec:Fault Chain Search Strategy}

We employ the standard $\epsilon$-greedy search strategy with an adaptive exploration schedule. Initially, the agent is compelled to take actions based on prior knowledge to find FCs with maximum expected $\tll$. Typically, since an outage of a component carrying higher power makes the remaining components vulnerable to overloading, a reasonable exploration strategy of the agent would be to remove components carrying maximum power-flow. Therefore, we follow a power-flow weighted $(\PFW)$ exploration strategy (also adopted in~\cite{Zhang:2020}) such that, in any stage $i$ of the FC, the agent chooses the $j^{\rm th}$ available component $\ell^{j}_{i} \in \mcA_{i}$ according to the rule
\begin{align}
    \label{eq:exploration policy}
    a_{i} = \argmax_{\ell^{j}_{i}} \; \frac{ \pf(\ell^{j}_{i})/\sqrt{{\sf count}(\bS_{i},\ell^{j}_{i}) + 1} }{\sum_{k=1}^{|\mcA_{i}|} \; \pf(\ell^{k}_{i})/\sqrt{{\sf count}(\bS_{i},\ell^{k}_{i}) + 1} }\ ,
\end{align} with probability $\epsilon$ where we denote $\pf(\ell^{j}_{i})$ as the~\emph{absolute} value of the power flowing through component $\ell^{j}_{i}$ and denote ${\sf count}(\bS_{i},\ell^{j}_{i})$ as the number of times the component $\ell^{j}_{i}$ was chosen when the agent was in~{POMDP} state $\bS_{i}$ in the past. On the other hand, as the agent gains more experience, the agent should choose actions based on the $Q$-values learned via the behaviour network $\GRQN(\cdot, \cdot | \btheta)$. Accordingly, a strategy based on $Q$-values learned by the agent is designed. Specifically, actions are chosen proportional to the $Q$-values normalized by each~{POMDP} state-action visit count to~\emph{avoid repetitions} of FC sequences discovered earlier. Accordingly, the agent chooses action $a_{i}$
\begin{align}
    \label{eq:exploitation policy}
    a_{i} = \argmax_{\ell^{j}_{i}} \; \frac{Q(\bY_{i}, \ell^{j}_{i} | \btheta)}{\sqrt{{\sf count}(\bS_{i}, \ell^{j}_{i}) + 1}}\ ,
\end{align} with probability $1 - \epsilon$.

In order to balance the exploration-exploitation trade-off between~\eqref{eq:exploration policy} and~\eqref{eq:exploitation policy} during the training of the $\GRQN$, it is important to dynamically alter the probability $\epsilon$ so that, with more experience, the agent chooses actions based on~\eqref{eq:exploitation policy}. Appropriately, we follow the exploration schedule given by 
\begin{align}
    \label{eq:epsilon update}
    \epsilon = \max\left(\frac{\sum_{j=1}^{|\mcA_{1}|} \; \pf(\ell^{j}_{1})/\sqrt{{\sf count}(\bS_{0},\ell^{j}_{1}) + 1}}{\sum_{k=1}^{|\mcA_{1}|} \; \pf(\ell^{k}_{1})},\; \epsilon_{0}\right)\ , 
\end{align} where $\epsilon_{0}$ ensures a minimum level of exploration.

\subsubsection{Offline Sequential Buffer Initialization}
\label{sec:Offline Sequential Buffer Initialization}

To initiate the parameter update of the behavior $\GRQN$ via gradient descent~\eqref{eq:gradient update}, there must exist at least $B$ FC sequences in the experience buffer. However, unlike in the case of~\eqref{eq:exploration policy}, the buffer can be populated~\emph{offline} for any loading condition. Therefore, the agent can afford to take actions greedily with respect to components conducting maximum power and, accordingly, backtrack to update the availability of actions to avoid repeating FCs discovered previously. Hence, prior to the start of our real-time FC search Algorithm~1, we let the agent explore~\emph{offline} for ${\sf Explore}$ iterations where the agent, in any stage $i$, chooses actions according to the rule
\begin{align}
    \label{eq:offline strategy}
    a_{i} = \argmax_{\ell^{j}_{i}} \; \frac{\pf(\ell^{j}_{i})}{\sum_{k=1}^{|\mcA_{i}|} \; \pf(\ell^{k}_{i})}\ ,
\end{align} with probability $1$ to fill the sequential experience buffer. Note that this needs to be done only once offline for any loading condition. This is important since the quality of the sequences in the buffer greatly affects the efficiency of the search.

\subsection{IEEE-39 New England Test System}
\label{sec:IEEE-39 New England Test System}

This test system comprises of $N = 39$ buses and $|\mcU| = 46$ components, including $12$ transformers and $34$ lines. We consider a loading condition of $0.55\times{\sf base\_load}$, where ${\sf base\_load}$ denotes the standard load data for the New England test case in PYPOWER~\emph{after} generation-load balance to quantify the performance of our approach. These loading conditions were chosen since it is relatively difficult to discover FCs with large $\tll$s in a lightly loaded power system as there are fewer such FCs in comparison to the space of all FC sequences $|{\mcF}|$.

\subsubsection{Parameters and Hyper-parameters} 
\label{sec:Parameters and Hyper-parameters}

The hyper-parameters of the $\GRQNs$ are chosen by performing hyper-parameter tuning. Accordingly, we choose $H = G = 12$ latent and output number of features when computing both the latent system state~\eqref{eq:GRNN time varying latent state} and the output system state~\eqref{eq:GRNN output}. We use $K = 3$ graph-shift operations for the graph-filter~\eqref{eq:graph filter} that is used to compute~\eqref{eq:GRNN time varying latent state} and~\eqref{eq:GRNN output}. We use both $\rho$ and $\sigma$ as the hyperbolic tangent non-linearity $\sigma = \rho = \tanh$ and the ${\sf ReLU}$ non-linearity for the fully-connected $\nn$ that approximates the $Q$-values. For other parameters, we choose $F=1$ since we employ voltage phase angles as the only input system state parameter $f$, choose $\gamma = 0.99$ since large $\loss$s mostly occur in the last few stages of the FC sequence, an $\epsilon_{0} = 0.01$ to ensure a minimum level of exploration during the FC search process, a batch size $B = 32$, ${\sf Explore} = 250$, a risk assessment horizon $P=3$, FC sequence iteration $S = 1200$ (\emph{excluding} the initial ${\sf Explore}$ iterations), learning rate $\alpha = 0.005$, and $\kappa \in [3]$ that controls the frequency of the learning update~\eqref{eq:gradient update} and also governs the computational complexity of the graph recurrent $Q$-learning algorithm. 

To quantify the regret, we need to compute the $\tll$ associated with the $S$ most critical FC sequences (ground truth) $\mcV^{*}_{s}$, $\forall s \in [S]$. This is carried out by generating~\emph{all possible} FC sequences (i.e., set $\mcF$) with a target horizon of $P=3$ for the considered total load of $0.55\times{\sf base\_load}$. By leveraging the pre-computed set $\mcF$, we observe a total of $3738$ risky FCs for the loading condition $0.55\times{\sf base\_load}$ using our developed FC simulator.

\subsection{Baseline Agents}
\label{sec:Baseline Agents}

To further illustrate the merits of the proposed graphical framework, we compare it with two state-of-the-art baseline approaches proposed in~\cite{Zhang:2020}. We label the first approach in~\cite{Zhang:2020} based on the ordinary $Q$-learning algorithm without prior knowledge as $\PFW$ + $\rl$ and label their best performing approach based on transition and extension of prior knowledge from other power system snapshots by $\PFW + \rl + \TE$. To ensure a fair comparison, we employ the~\emph{same} exploration schedule for $\epsilon$ discussed in Section~\ref{sec:Fault Chain Search Strategy} with the same parameters and the same discount factor $\gamma$ for all the approaches. Note that, in the $\PFW+\rl+\TE$ approach, we first run their proposed $Q$-learning based approach offline, for a loading condition of $0.6\times{\sf base\_load}$ (bringing in the prior knowledge). This is run for $S = 5000$ iterations to ensure the convergence of their $Q$-learning algorithm. Subsequently, we store its extensive $Q$-table to run its $\PFW+\rl+\TE$ approach in real-time for the considered loading condition of $0.55\times{\sf base\_load}$, signifying a transition from the power system snapshot loaded at $0.6\times{\sf base\_load}$ and an extension to the current power system snapshot loaded at $0.55\times{\sf base\_load}$. Note that, when performing comparisons, we set the parameters and hyper-parameters associated with Algorithm~1 the same as that described in Section~\ref{sec:Parameters and Hyper-parameters}.

\subsection{IEEE-118 Test System}
\label{sec:IEEE-118 test system}

This test system consists of $N = 118$ buses and $|\mcU| = 179$ components. We consider a loading condition of $0.6 \times {\sf base\_load}$, where ${\sf base\_load}$ denotes the standard load data for the IEEE-$118$ test case in PYPOWER after generation-load balance to quantify the performance of our approach.

\subsubsection{Parameters and Hyper-parameters}
\label{sec:Parameters and Hyper-parameters 118}

{We choose $H = G = 48$ latent and output number of features when computing both the latent system state~\eqref{eq:GRNN time varying latent state} and the output system state~\eqref{eq:GRNN output}. We use $K = 3$ graph-shift operations for the graph-filter~\eqref{eq:graph filter}, use both $\rho$ and $\sigma$ as the hyperbolic tangent non-linearity $\sigma = \rho = \tanh$ and the ${\sf ReLU}$ non-linearity for the fully-connected $\nn$ to approximate $Q$-values. For other parameters, we choose $F=1$ since we employ voltage phase angles as the only input system state parameter $f$, choose $\gamma = 0.99$, $\epsilon_{0} = 0.01$, a batch size $B = 32$, ${\sf Explore} = 250$, a risk assessment horizon $P=3$, FC sequence iteration $S = 1600$ (\emph{excluding} the initial ${\sf Explore}$ iterations), learning rate $\alpha = 0.0005$, and $\kappa \in [3]$.

\subsubsection{Accuracy and Efficiency -- Performance Results}
\label{sec:Accuracy and Efficiency - Performance Results 118}

{Algorithm~1 is initiated after the experience buffer is filled for ${\sf Explore}$ search iterations. Table~\ref{table:118 bus} illustrates the results obtained. Similar to $39$-bus system, we observe that Algorithm~1 with a greater $\kappa$ discovers FC sequences with larger accumulated $\tll$s and discovers more number of risky FCs leading to superior accuracy metrics for larger $\kappa$. For instance the average regret of Algorithm~1 with $\kappa = 3$ is $321.90 \times 10^{3}$ MWs and it is $0.4 \%$ lower the average regret when $\kappa = 1$. Similarly, the average precision when $\kappa = 3$ is $11.7\%$ higher that of $\kappa = 1$. Figure~\ref{fig:Accum Risk 118} illustrate the average accuracy metric for Algorithm~1 as a function of $s\in[S]$. A comprehensive discussion of additional performance results is provided in~\cite{TPS-Dwivedi:2024}.
}

\subsubsection{Comparison with Baselines} 
\label{sec:Comparison with Baselines 118}

{We perform comparisons with the two baselines approaches in~\cite{Zhang:2020}, i.e., the $\PFW$ + $\rl$ and $\PFW + \rl + \TE$. For the $\PFW+\rl+\TE$ approach, we first run their proposed $ Q$-learning-based approach offline, for a loading condition of $1.0\times{\sf base\_load}$ (bringing in the prior knowledge) for $S = 5000$ iterations and store its extensive $Q$-table to run the $\PFW+\rl+\TE$ approach in real-time for the considered loading condition of $0.6\times{\sf base\_load}$, signifying a transition from $1.0\times{\sf base\_load}$ to an extension to the current system loaded at $0.6\times{\sf base\_load}$. We set the parameters and hyper-parameters as described in Section~\ref{sec:Parameters and Hyper-parameters 118}.}

\paragraph{Comparison under Unbounded Computational Budget} Table~\ref{table:118 bus} compares the accuracy metrics for $S=1600$, showing that Algorithm~1 consistently outperforms both the other baseline approaches. Figure~\ref{fig:Accum Risk 118} shows how the average regret scales as a function of search iterations $s\in [S]$.

\paragraph{Comparison under Bounded Computational Budget} We next evaluate all the above approaches considering a run-time computational budget of five minutes, averaged over $25$ $\mc$ iterations, and Table~\ref{table:time_118 bus} illustrates the relative performance. All the observations corroborate those observed for the 39-bus system.

\vspace{0.3in}

\begin{figure}[h]
    \centering
    \includegraphics[width=0.7\linewidth]{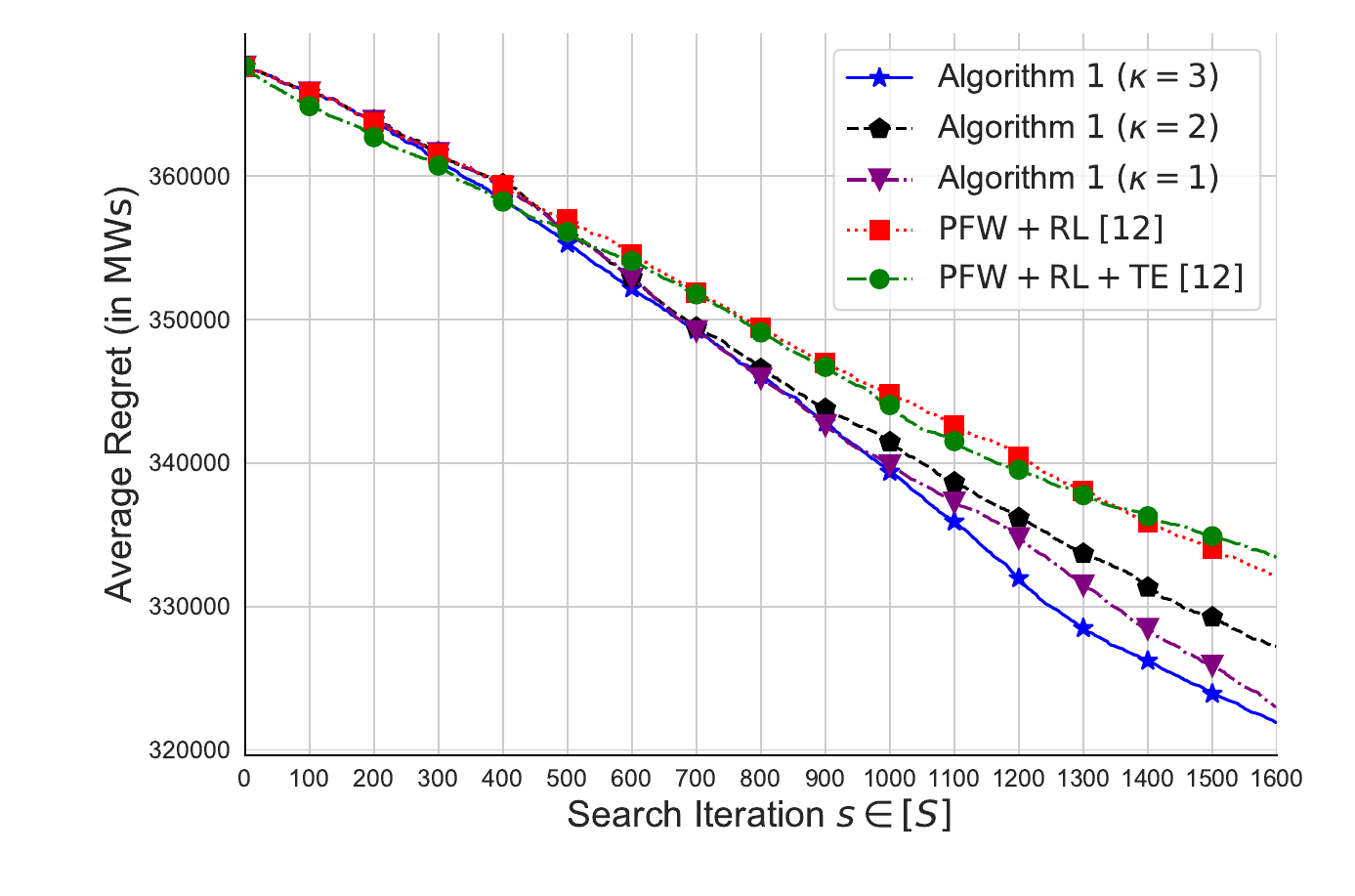}
    \vspace{-0.3in} 
    \caption{${\sf Regret}(s)$ versus $s$ for the IEEE-$118$ bus system.}
    \label{fig:Accum Risk 118}
\end{figure}

\vspace{0.3in}

\begin{table*}[h]
\centering
\scalebox{0.8}{
	\begin{tabular}{| c | c | c | c | c |} 			
		\hline
		\thead{Algorithm}               
		& \thead{Range for Accumulative $\tll$\\ $\sum_{i=1}^{S} \tll(\mcV_{i})$ (in GWs)}   
		& \thead{Range for} ${\sf Regret}(S)$ (in GWs)\\ [2.0ex] 
		\hline
		
		Algorithm~1$\;(\kappa = 3)$    & $45.73\;\pm\;16\%$   & $321.90\;\pm\;2.3\%$    \\ [1.0ex] 
		\hline
		
		Algorithm~1$\;(\kappa = 2)$    & $40.42\;\pm\;17\%$    & $327.22\;\pm\;2.4\%$      \\ [1.0ex] 
		\hline
		
		Algorithm~1$\;(\kappa = 1)$    & $44.62\;\pm\;12\%$     & $323.02\;\pm\;1.7\%$    \\ [1.0ex] 
		\hline
	    
	    $\PFW$ + $\rl$ + $\TE$ \cite{Zhang:2020} & $34.23\;\pm\;2.7\%$   & $333.40\;\pm\;0.28\%$    \\[1.0ex] 
		\hline
		
		$\PFW$ + $\rl$ \cite{Zhang:2020}  & $35.50\;\pm\;2.4\%$   & $332.14\;\pm\;0.25\%$     \\ [1.0ex] 
		\hline
	\end{tabular}%
	}
	\caption{{Performance comparison for the IEEE-$118$ test system.}}
	\label{table:118 bus}
\end{table*}

\vspace{0.3in}

\begin{table*}[h]
\centering
\scalebox{0.8}{
	\begin{tabular}{| c | c | c | c | c | c |} 			
		\hline
		\thead{Algorithm}     
		& \thead{Average No. of FC \\ Sequences $S$ Discovered}
		& \thead{Range for Accumulative $\tll$\\ $\sum_{i=1}^{S} \tll(\mcV_{i})$ (in GWs)}   
		& \thead{Range for} ${\sf Regret}(S)$ (in GWs) \\ [2.0ex] 
		\hline
		
		Algorithm~1$\;(\kappa = 3)$  & $186$  & $11.46\;\pm\;16.2\%$  & $175.63\;\pm\;3.4\%$   \\ [1.0ex] 
		\hline
		
		Algorithm~1$\;(\kappa = 2)$  & $239$  & $18.32\;\pm\;18\%$    & $212.22\;\pm\;4.1\%$   \\ [1.0ex] 
		\hline
		
		Algorithm~1$\;(\kappa = 1)$  & $301$ & $19.86\;\pm\;13\%$     & $256.75\;\pm\;2.6\%$   \\ [1.0ex] 
		\hline
	    
	    $\PFW$ + $\rl$ + $\TE$ \cite{Zhang:2020} & $527$ & $18.56\;\pm\;3\%$   & $436.22\;\pm\;0.98\%$  \\[1.0ex] 
		\hline
		
		$\PFW$ + $\rl$ \cite{Zhang:2020} & $522$  & $17.98\;\pm\;3\%$   & $439.13\;\pm\;1.1\%$   \\ [1.0ex] 
		\hline
	\end{tabular}%
	}
	\caption{{Performance comparison for a computational time of $5$ minutes for the IEEE-$118$ bus test system.}}
	\label{table:time_118 bus}
\end{table*}

\end{document}